\documentclass{IET}

\usepackage{algorithmic}
\usepackage{array}

\begin{document}

\title{Learning Discriminative Features using Encoder/Decoder type Deep Neural Nets}

\author{Vishwajeet Singh}
\affil{ALPES, Bolarum, Hyderabad 500010, vsthakur@gmail.com}

\author{Killamsetti Ravi Kumar}
\affil{ALPES, Bolarum, Hyderabad 500010, ravi.killamsetti@gmail.com}

\author{K Eswaran}
\affil{SNIST, Ghatkesar, Hyderabad 501301, kumar.e@gmail.com}
\abstract{As machine learning is applied to an increasing variety of complex problems, which are defined by high dimensional and complex data sets, the necessity for ``task oriented feature learning'' grows in importance. With the advancement of Deep Learning algorithms, various successful feature learning techniques have evolved.

In this paper, we present a novel way of learning discriminative features by training Deep Neural Nets which have Encoder/Decoder type architecture similar to an Autoencoder. We demonstrate that our approach can learn discriminative features which can perform better at pattern classification tasks when the number of training samples is relatively small in size.}

\maketitle

\section{Introduction}
In the field of machine learning and statistics, many linear (\cite{pearson1901}), nonlinear (\cite{SchoelkopfEtAl:1999} \& \cite{Roweis00nonlineardimensionality}) and stochastic (\cite{Bingham:2001:RPD:502512.502546}) methods have been developed to reduce the dimensionality of data so that relevant information can be used for classification of patterns (\cite{joachims1998text} \& \cite{journals/corr/abs-1207-3538}). Researchers have solved pattern recognition problems (to varying degrees of success) like face detection \cite{journals/pami/GarciaD04}, gender classification \cite{journals/tnn/PhungB07}, human expression recognition \cite{journals/tnn/RosenblumYD96}, object learning \cite{journals/nn/BaldiH89}, unsupervised learning of new tasks \cite{citeulike:2893893} and also have studied complex neuronal properties of higher cortical areas \cite{LauStanleyDan02}. However, most of the above techniques did not require automatic feature extraction as a pre-processing step to pattern classification.

\begin{figure}[h]
\begin{center}
\includegraphics[scale=0.5]{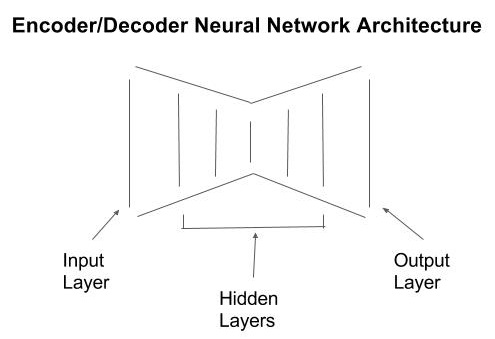}
\caption{Architecture of Encoder/Decoder Type Neural Networks}
\label{fig:encoder-decoder-arch}
\end{center}
\end{figure}

In contrast to the above, there exist many practical applications characterized by high dimensionality of
data (such as speech recognition, remote sensing, e.t.c), where finding sufficient labeled examples
might not be affordable or feasible. At the same time there may be lot of unlabeled data available
easily. Unsupervised feature learning techniques, like the Autoencoder (\cite{HintonSalakhutdinov2006b}, \cite{journals/jmlr/SalakhutdinovH07} , \cite{journals/corr/abs-0911-0225} and \cite{DBLP:journals/corr/abs-0812-2535}), try to capture the essential
structure underlying the high-dimensional input data by converting them into lower dimensional data
without losing information. Autoencoder follows an Encoder/Decoder type neural network architecture (see figure \ref{fig:encoder-decoder-arch}), where the dimensionality of the input and the output layers are the same. The output of the network is forced (via learning) to be the same as it's input. Typically all the other layers in the network are smaller in size when dimensionality reduction is the goal of learning. This way they learn features that are much lower in dimension as compared to the input data and are rich in information to later perform pattern classification on the labeled data sets.

The primary aim of dimensionality reduction for pattern classification problems is to remove the unnecessary information from data and extract information which is meaningful for achieving efficient pattern recognition/classification. With the advent of Autoencoder and various forms of Unsupervised Feature Learning, a significant amount of success is achieved in this aspect. But these methods demand large amount of data to be available for learning.

\begin{figure}[h]
\begin{center}
\includegraphics[scale=0.50]{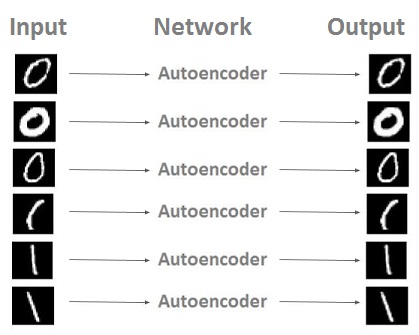}
\caption{Input-to-Output Mapping of an Autoencoder}
\label{fig:ae-mapping}
\end{center}
\end{figure}

Another very important aspect is that by mapping the input back to itself as output, the Autoencoder network retains lot of additional information present in the input which is not relevant to the pattern classification problem. To elaborate further, figure \ref{fig:ae-mapping} depicts the mapping of an Autoencoder where it is trying to learn handwritten digits. The first two inputs, although they represent the same character zero, the network is forced to learn the thickness and the exact shape of the handwritten digit. Features learnt by this approach still contain lot of information which is not useful for pattern classification and hence can be treated as noisy. When the amount of data available to train these networks is reduced, their ability to learn discriminative features also reduces significantly, as will be shown in section \ref{comp-analysis-sec}.

In this paper we focus on the scenario where there is very little labeled data per class and zero unlabeled data available. In this context we  describe a novel way of learning discriminative features using Deep Neural Nets which have an Encoder/Decoder architecture (see figure \ref{fig:encoder-decoder-arch}). We refer to this network as ``Discriminative Encoder''. Section \ref{discriminative-encoder-sec} introduces the concept of ``Discriminative Encoder'' and explains how it is different from the Autoencoder. Sections \ref{experiments-sec} and \ref{comp-analysis-sec} provide the results of benchmarking ``Discriminative Encoder'' on standard machine learning data sets. The unique feature of this study is that we have benchmarked the performance on data sets of varying sizes in terms of number of training samples and number of classes. Lastly, Section \ref{conclusion-sec} concludes with the findings and future direction.

\section{Discriminative Encoder}
\label{discriminative-encoder-sec}

The motivation behind this approach is to extract meaningful information from a relatively small set of labeled samples such that:
\begin{enumerate}
\item  features learnt are less sensitive to intra-class difference in the inputs of samples belonging to the same class
\item  features learnt are highly sensitive to inter-class differences in the inputs of samples belonging to different class
\end{enumerate}

\begin{figure}[h]
\begin{center}
\includegraphics[scale=0.50]{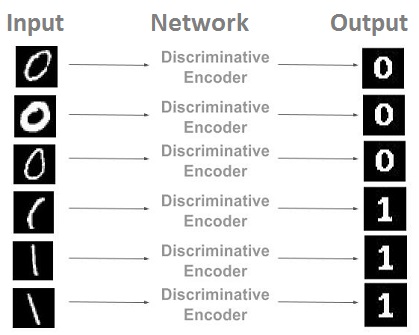}
\caption{Input-to-Output Mapping of a Discriminative Encoder}
\label{fig:de-mapping}
\end{center}
\end{figure}

To achieve this we use the Encoder/Decoder neural network architecture similar to the Autoencoder. The difference is that instead of mapping the input back to itself as output (i.e, output = input), we make the input map to an ``Ideal Input'' of the class. Figure \ref{fig:de-mapping} depicts this concept, here we are making all the handwritten zeros to map to an ``Ideal Zero'' (or a standard template of zero). Similarly, all the handwritten ones are mapped to an ``'Ideal One''.

This mapping forces the network to ignore the differences between samples belonging to the same class and focus on differences between samples belonging to different classes. In simple words, the features learnt this way are ``discriminative'' in nature.

\section{Experiments}
\label{experiments-sec}
The purpose of the present exercise is to benchmark the Discriminative Encoder network on datasets of varying sizes in terms of the number of training samples and the number of classes. Table \ref{table:dataset-summary} summarizes the datasets used in this study.

\begin{table}[h]
\begin{center}
\begin{tabular}{| >{\centering\arraybackslash}m{2.5cm} | >{\centering\arraybackslash}m{2.5cm} | >{\centering\arraybackslash}m{2.5cm} | >{\centering\arraybackslash}m{2.5cm} |}
\hline
Name & Number of classes & Total number of sample & \#Input Features\\
\hline
Extended Yale Face Dataset (Frontal Pose) & 38 & 2432 & 900 \\ \hline
Extended Yale Face Dataset (All Poses) & 28 & 11482 & 900 \\ \hline
NCKU Taiwan Face Dataset & 90 & 3330 & 768 \\ \hline
MNIST Dataset & 10 & 70000 & 784 \\ \hline
\end{tabular}
 \\
\end{center}
\caption{List of datasets used for benchmarking}
\label{table:dataset-summary}
\end{table}

The uniqueness of this work is that we have used much compact or simpler models,  in terms of number of parameters,  when compared to similar work in the field of Deep Learning (\cite{bib1}). The results obtained are comparable with the state-of-the research in deep learning (some of which are cited).

\subsection{Extended Yale B Data Set (Frontal Pose)}
\begin{figure}[h]
\begin{center}
\includegraphics[scale=0.40]{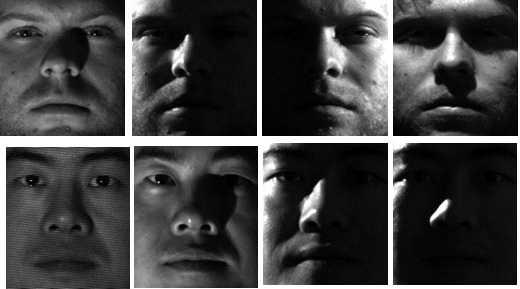}
\caption{Sample faces from Extended Yale B (Frontal Pose) dataset}
\label{fig:yale-sample-images}
\end{center}
\end{figure}

The Extended Yale B (Frontal Pose) Face data set (\cite{journals/pami/GeorghiadesBK01} and \cite{journals/pami/LeeHK05}) consists of frontal images of 38 subjects taken across 64 different illumination conditions (64 images per subject and 2432 images in total). The size of the original images was 168x192, which was reduced to 30x30 in our experimentation (i.e. we have 900 features per input image). Some of the sample images from this data set are shown in figure \ref{fig:yale-sample-images}.

To train the Discriminative Encoder of dimension $400-200-64-900$ (this convention represent the number of processing elements in each layer), 48 images per subject (1824 in total) were taken from the data set and the remaining 16 per subject (608 in total) were kept for testing the classifier. Once the network was trained with high accuracy, the data set was reduced from 900 dimension feature vector to 64 dimensional feature vector. The results of using supervised classifiers on the 64 dimensional data set are described in table \ref{table:cropped-yale-res}

\begin{figure}[h]
\begin{center}
\includegraphics[scale=0.40]{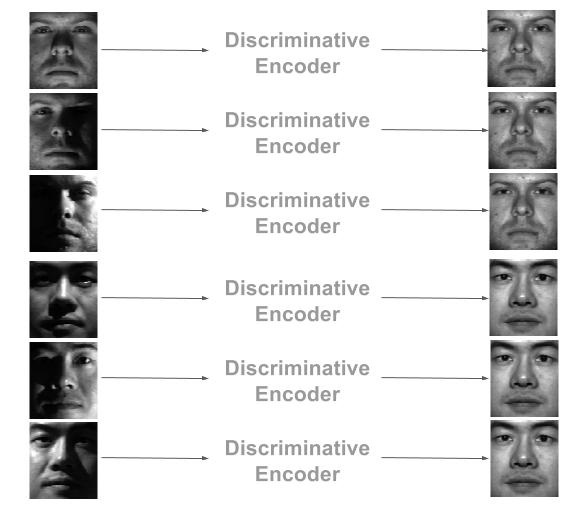}
\caption{Input-to-Output Mapping of a Discriminative Encoder for the Extended Yale B (Frontal Pose) dataset}
\label{fig:yale-de-mapping}
\end{center}
\end{figure}

At this point, we would like to highlight that this particular data set has been used extensively in the Machine Learning community to benchmark Face Recognition algorithms, although a lot of the published work makes use of domain specific knowledge to build their face recognition systems and can possibly ignore the dark images in the dataset (see table \ref{table:yale-frontal-compare} for comparison). The work in \cite{conf/icml/TangSH12a} has utilized large databases for 3D morphological modeling and albedo modeling. We have neither made use of any domain specific knowledge, nor have we removed any dark images from the data set. It may be noticed that classification in the reduced dimensional space, in general, would give a better classification as the noise in the original data set would have been removed during training.

\begin{table}[h]
\begin{center}
\begin{tabular}{| >{\centering\arraybackslash}m{2cm} | >{\centering\arraybackslash}m{2cm} | >{\centering\arraybackslash}m{2cm} | >{\centering\arraybackslash}m{2cm} |}
\hline
Classifier & Setting & Accuracy in $R^{900}$ & Accuracy in $R^{64}$\\
\hline
Neural Network & 75-50-38 & & 98.3\% \\ \hline
k-Nearest Neighbor & k=3 & 60.6\% & 97.3\% \\ \hline
k-Nearest Neighbor & k=5 & 60.3\% & 97.5\% \\ \hline
k-Nearest Neighbor & k=7 & 58.5\% & 97.5\% \\ \hline
k-Nearest Neighbor & k=9 & 56.7\% & 97.5\% \\ \hline
\end{tabular}
\\
\end{center}
\caption{Results on Extended Yale B (Frontal) Face data set}
\label{table:cropped-yale-res}
\end{table}

\begin{table}[h]
\begin{center}
\begin{tabular}{| >{\centering\arraybackslash}m{2cm} | >{\centering\arraybackslash}m{2cm} | >{\centering\arraybackslash}m{2cm} | >{\centering\arraybackslash}m{2cm} | >{\centering\arraybackslash}m{2cm} | }
\hline
Study & \#Subjects & \#Train Images per Subject & \#Model Params (million) & Accuracy\\
\hline
Current Paper & 38 & 48  & 0.5 & 98.3\% \\ \hline
Hinton et. al. \cite{conf/icml/TangSH12a} & 10 & 7 & 1.3 & 97\% \\ \hline
\end{tabular}
\\
\end{center}
\caption{Comparison of results on Extended Yale B (Frontal) data set}
\label{table:yale-frontal-compare}
\end{table}

\subsection{Extended Yale B Data Set (All Poses)}
\begin{figure}[h]
\begin{center}
\includegraphics[scale=0.35]{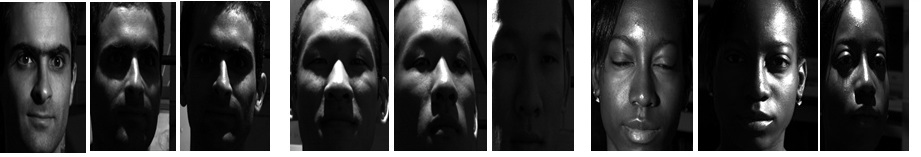}
\caption{Sample of faces from Extended Yale B (All Pose) Face data set}
\label{fig:yale-all-pose-sample-images}
\end{center}
\end{figure}

The Extended Yale B data set (\cite{journals/pami/GeorghiadesBK01}) consists of images of 28 subjects taken across 9 poses and 64 different illumination conditions (576 images per subject and 16128 images in total). Some of the sample images from this data set are shown in figure \ref{fig:yale-all-pose-sample-images}.

The original images contained lot of background information and hence we had to extract only the faces from the images first, which was done using OpenCV library. Of the total 16128 images, faces could be detected only in 11482 images and the rest were quiet dark for the faces to be detected. The reduced data set contains approximately 410 images per subject. The size of the images was reduced to 30x30 in our experimentation (i.e. we have 900 features per input image).

\begin{table}[h]
\begin{center}
\begin{tabular}{| >{\centering\arraybackslash}m{2cm} | >{\centering\arraybackslash}m{2cm} | >{\centering\arraybackslash}m{2cm} | >{\centering\arraybackslash}m{2cm} |}
\hline
Classifier & Setting & Accuracy in $R^{900}$ & Accuracy in $R^{64}$\\
\hline
Neural Network & 75-50-38 & & 95.7\% \\ \hline
k-Nearest Neighbor & k=3 & 81.6\% & 95.4\% \\ \hline
k-Nearest Neighbor & k=5 & 81.3\% & 95.4\% \\ \hline
k-Nearest Neighbor & k=7 & 81.0\% & 95.4\% \\ \hline
k-Nearest Neighbor & k=9 & 80.5\% & 95.3\% \\ \hline
\end{tabular}
\\
\end{center}
\caption{Results on Extended Yale B (All Pose) face data set}
\label{table:yale-all-pose-res}
\end{table}

To train the Discriminative Encoder of dimension $400-200-64-900$, 8600 images were taken from the reduced data set and the remaining 2882 images were kept for testing the classifier. Once the network was trained with high accuracy, the data set was reduced from 900 dimension feature vector to a 64 dimensional feature vector. The results of using supervised classifiers on the 64 dimensional data set are described in table \ref{table:yale-all-pose-res}

\subsection{Taiwan Face Data Set}
\begin{figure}[h]
\begin{center}
\includegraphics[scale=0.35]{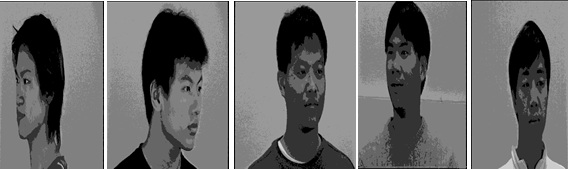}
\caption{Sample of faces from Taiwan Face data set}\label{fig:taiwan-sample-images}
\end{center}
\end{figure}

This data set \cite{TaiwanDatabase} is provided by the Robotics Lab of the Dept of Computer Science of National Cheng Kung University in Taiwan. The whole database contains 6660 images of 90 subjects. Each subject has 74 images, where 37 images were taken every 5 degree from right profile (defined as $+90^o$) to left profile (defined as $-90^o$) in the pan rotation. The remaining 37 images are generated (synthesized) by the existing 37 images using commercial image processing software in the way of flipping them horizontally. Some sample images from the dataset are shown in Figure \ref{fig:taiwan-sample-images}

In our experiments, we have considered only half of this data set, i.e., 3330 images of 90 subjects and each subject has 37 images which were taken every 5 degree from right profile (defined as $+90^o$) to left profile (defined as $-90^o$) in the pan rotation. In all the images, only the face part of the image was retained and the region containing the clothes on subjects body were trimmed from the original image. Later the images were reduced to 24x32 pixels size (i.e. 768 features).

\begin{table}[h]
\begin{center}
\begin{tabular}{| >{\centering\arraybackslash}m{2cm} | >{\centering\arraybackslash}m{2cm} | >{\centering\arraybackslash}m{2cm} | >{\centering\arraybackslash}m{2cm} |}
\hline
Classifier & Setting & Accuracy in $R^{768}$ & Accuracy in $R^{25}$\\
\hline
Neural Network & 25-50-90 & & 99.5\% \\ \hline
k-Nearest Neighbor & k=3 & 97.171\% & 99.6\% \\ \hline
k-Nearest Neighbor & k=5 & 94.44\% & 99.6\% \\ \hline
k-Nearest Neighbor & k=7 & 91.81\% & 99.6\% \\ \hline
k-Nearest Neighbor & k=9 & 89.09\% & 99.6\% \\ \hline
\end{tabular}
\\
\end{center}
\caption{Results on Taiwan Face data set}
\label{table:taiwan-res}
\end{table}

To train network of dimension $196-64-25-768$, 26 images per subject (2340 in total) were taken from the data set and the remaining 11 per subject (990 in total) were kept for testing the classifier. Once the network was trained, the data set was reduced from 768 dimension feature vector to a 25 dimensional feature vector. The results of using supervised classifiers on the 25 dimensional data set are described in table \ref{table:taiwan-res}

\subsection{MNIST Data Set}

The MNIST database (\cite{lecun-gradientbased-learning-applied-1998}) of images of handwritten digits (0-9) is a standard benchmark data set used in the machine learning community. It has a training set of 60,000 examples (approximately 6000 examples per digit), and a test set of 10,000 examples. The dimensionality of images is 28x28 (i.e. 784 features per input to the network).

\begin{figure}[h]
\begin{center}
\includegraphics[scale=0.45]{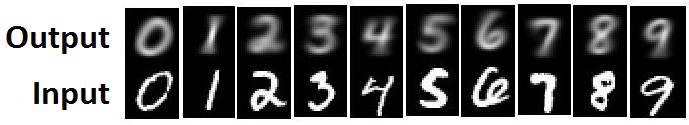}
\caption{MNIST Input and Output of the trained network}\label{fig:mnist-pred}
\end{center}
\end{figure}

The network architecture contains $225-100-36-784$ processing elements in the layers of the network. With sufficient amount of training, this network was able to learn all the mapping with high accuracy (See Figure \ref{fig:mnist-pred}). The trained network was then used to reduce the dimensionality of the entire data set from $R^{784}$ to $R^{36}$. Table \ref{table:mnist-res} shows the results of using supervised classifiers (k-Nearest Neighbor classifier and Neural Network) to classify the reduced $36$ dimensional data.

\begin{table}[h]
\begin{center}
\begin{tabular}{| >{\centering\arraybackslash}m{2cm} | >{\centering\arraybackslash}m{2cm} | >{\centering\arraybackslash}m{2cm} | >{\centering\arraybackslash}m{2cm} |}
\hline
Classifier & Setting & Accuracy in $R^{784}$ & Accuracy in $R^{36}$\\
\hline
Neural Network & 36-5-10 & & 98.08\% \\ \hline
k-Nearest Neighbor & k=3 & 97.05\% & 97.5\% \\ \hline
k-Nearest Neighbor & k=5 & 96.88\% & 97.5\% \\ \hline
k-Nearest Neighbor & k=7 & 96.94\% & 97.6\% \\ \hline
k-Nearest Neighbor & k=9 & 96.59\% & 97.6\% \\ \hline
\end{tabular}
\\
\end{center}
\caption{Results on MNIST data set}
\label{table:mnist-res}
\end{table}

In the case of MNIST data set, k-Nearest Neighbor works in the high dimensional space due to the availability of lot of training data, which appears to be reasonably clustered.

Some of the state-of-the-art algorithms, like \cite{journals/jmlr/SalakhutdinovH07} and \cite{journals/corr/abs-1003-0358}, use atleast $7$ times more the number of parameters (weights) as compared to the ones used in this paper (see table \ref{table:mnist-compare}).

\begin{table}[h]
\begin{center}
\begin{tabular}{| >{\centering\arraybackslash}m{2cm} | >{\centering\arraybackslash}m{2.5cm} | >{\centering\arraybackslash}m{2cm} | >{\centering\arraybackslash}m{2cm} |}
\hline
Study & Method & \#Model Params (million) & Accuracy\\
\hline
This Paper & Discriminative Encoder  & 0.23 & 98.08\% \\ \hline
Hinton et. al. \cite{journals/jmlr/SalakhutdinovH07} & Autoencoder & 1.7 & 99\% \\ \hline
Schmidhuber et. al. \cite{journals/corr/abs-1003-0358} & Simple Deep Neural Nets + Elastic Distortions & 11.9 mil & 99.65\% \\ \hline
\end{tabular}
\\
\end{center}
\caption{Comparison of results on MNIST data set}
\label{table:mnist-compare}
\end{table}

\section{Comparative Analysis}
\label{comp-analysis-sec}

This section discusses the results of comparative analysis of a k-Nearest Neighbor (kNN) classifier, here k=3, performance on various dimensionality reduction approaches. Table \ref{table:summary-res} shows the results of performing kNN classification on the data sets in the original input space (IS), after dimensionality reduction by principal component analysis (PCA), after dimensionality reduction by Autoencoder (AE) and finally after dimensionality reduction by Discriminative Encoder (DE). The table also shows the network architectures of Autoencoder and Discriminative Encoder. It is also important to note that we have not used Boltzman pre-training for either Autoencoder or for Discriminative Encoder. Backpropagation algorithm with mini-batch gradient descent was used to train the networks after random initialization of weights.

\begin{table}[h]
\begin{center}
\begin{tabular}{| >{\centering\arraybackslash}m{2cm} | >{\centering\arraybackslash}m{1.5cm} | >{\centering\arraybackslash}m{1.5cm} | >{\centering\arraybackslash}m{2cm} | >{\centering\arraybackslash}m{2cm} | >{\centering\arraybackslash}m{1cm} | >{\centering\arraybackslash}m{1cm} | >{\centering\arraybackslash}m{1cm} | >{\centering\arraybackslash}m{1cm} |}
\hline
Dataset & Input Space Size & Reduced Space Size & Network (AE) & Network (DE) &  IS & PCA & AE & DE\\ \hline
Yale (Frontal Pose) & 900 & 64 & 400-200-64-200-400-900 & 400-200-64-900 & 60.6\% & 51.4\% & 82.4\% & 97.3\% \\ \hline
Yale (All Poses) & 900 & 64 & 400-200-64-200-400-900 & 400-200-64-900 & 81.6\% & 74.6\% & 89.1\% & 95.4\% \\ \hline
Taiwan Face Db & 768 & 25 & 196-64-25-64-196-768 & 196-64-25-768 & 97.1\% & 96.9\% & 96.8\% & 99.6\% \\ \hline
MNIST & 784 & 36 & 225-100-36-100-225-784 & 225-100-36-784 & 97.0\% & 97.3\% & 97.0\% & 97.5\% \\ \hline
\end{tabular}
\\
\end{center}
\caption{Results of 3-NN classifier on all datasets using various dimensionality reduction approaches: IS (original input space), PCA (principal component analysis), AE (autoencoder), DE (discriminative encoder)}
\label{table:summary-res}
\end{table}

\begin{itemize}

\item From tables \ref{table:dataset-summary} and \ref{table:summary-res}, we can see that the ``Discriminative Encoder'' very clearly outperforms PCA and Autoencoder on Extended Yale (Frontal Pose) Face dataset where the the number of samples is the least. It also performs much better on Extended Yale (All Pose) Face dataset and on Taiwan Face dataset as compared to PCA and Autoencoder. When the number of samples increase in the MNIST case, we can see that the performance of all the dimensionality reduction approaches (PCA, Autoencoder and Discriminative Encoder) is almost alike. These results support our claim that the Discriminative Encoder is good at extracting discriminative features even when the number of samples is less.

\item An observation regarding the performance of Autoencoder and Discriminative Encoder on the Yale dataset. It can be seen that the performance of Autoencoder increases in ``All Pose'' dataset when compared to ``Frontal Pose'' dataset, while the performance of Discriminative Encoder decreases. Autoencoders improved performance can directly be attributed to the increase in the availability of training data. In case of Discriminative Encoder, the slight decrease in performance is due to the fact that the mapping that it is trying to learn is getting complicated, wherein the network tries to map different poses and illumination conditions to the frontal pose and illumination condition. Overall, the Discriminative Encoder performs much better than the Autoencoder on both of these datasets.

\item How does Discriminative Encoder perform better when there are few training samples ?  Discriminative Encoders forces all the samples belonging to the same class map to the ``Ideal Input'' of that class. This is a kind of supervisory feedback in the learning process, which the Autoencoder does not have. Due to this supervisory feedback the Discriminative Encoder receives during the training, it is able to extract lot of discriminative information available in the training set.
\end{itemize}

\section{Conclusion}
\label{conclusion-sec}
In this paper, we have presented a novel way of learning discriminative features by training Encoder/Decoder type Deep Neural Nets. We have demonstrated that our approach can learn discriminative features which can perform better at pattern classification tasks when the number of training samples is relatively small in size. Also, we have found that when the number of samples to train are less in number, then relatively smaller sized networks (fewer processing elements per layer) can learn complex features, without any domain specific knowledge, and give high performance on pattern recognition tasks.

We would like to further our research by introducing the stacking and denoising approaches to train deep neural networks (\cite{Vincent:2010:SDA:1756006.1953039}). Also we would like to explore feature learning in an semi-supervised setting.

\end{document}